\documentclass[10pt, a4paper]{article}
\usepackage{lrec-coling2024} 

\usepackage{multirow}
\usepackage{amsmath}
\usepackage{algorithm}
\usepackage{algorithmic}

\title{\textbf{Few-shot Link Prediction on Hyper-relational Facts}}

\name{Jiyao Wei$^{1,2}$, Saiping Guan$^{2}$\sthanks{\ \ Corresponding authors}, Xiaolong Jin$^{1,2}$$^{*}$, Jiafeng Guo$^{1,2}$, Xueqi Cheng$^{1,2}$}

\address{$^{1}$School of Computer Science and Technology, University of Chinese Academy of Sciences;\\ $^{2}$Key Laboratory of Network Data Science and Technology,\\ Institute of Computing Technology, Chinese Academy of Sciences. \\
          \{weijiyao20z, guansaiping, jinxiaolong, guojiafeng, cxq\}@ict.ac.cn\\}

\abstract{
  Hyper-relational facts, which consist of a primary triple (head entity, relation, tail entity) and auxiliary attribute-value pairs, are widely present in real-world Knowledge Graphs (KGs). Link Prediction on Hyper-relational Facts (LPHFs) is to predict a missing element in a hyper-relational fact, which helps populate and enrich KGs. However, existing LPHFs studies usually require an amount of high-quality data. They overlook few-shot relations, which have limited instances, yet are common in real-world scenarios. Thus, we introduce a new task, Few-Shot Link Prediction on Hyper-relational Facts (FSLPHFs). It aims to predict a missing entity in a hyper-relational fact with limited support instances. To tackle FSLPHFs, we propose MetaRH, a model that learns Meta Relational information in Hyper-relational facts. MetaRH comprises three modules: relation learning, support-specific adjustment, and query inference. By capturing meta relational information from limited support instances, MetaRH can accurately predict the missing entity in a query. As there is no existing dataset available for this new task, we construct three datasets to validate the effectiveness of MetaRH. Experimental results on these datasets demonstrate that MetaRH significantly outperforms existing representative models.
 \\ \newline  }

\begin{document}

\maketitleabstract

\section{Introduction}
Link prediction aims to predict a missing element in an incomplete link within KGs. It plays a crucial role in enriching KGs and improving the performance of downstream applications like Web search and question answering~\cite{DBLP:conf/acl/DongWZX15,DBLP:conf/www/LukovnikovFLA17}. Previous research primarily focuses on binary facts, which are represented as triples (head entity, relation, tail entity). However, real-world KGs often contain hyper-relational facts that involve two entities and several auxiliary attribute-value pairs~\cite{Codd}. For instance, more than a third of entities in the popular KG Freebase~\cite{bollacker2008freebase} are involved in hyper-relational facts~\cite{DBLP:conf/ijcai/WenLMCZ16}. Therefore, it is essential to extend link prediction beyond binary facts.

In previous approaches to Link Prediction on Hyper-relational Facts (LPHFs), a hyper-relational fact is decomposed into multiple binary facts using virtual entities~\cite{nguyen2014don, krieger2015extending}. However, this decomposition results in the loss of structure information and increases the number of required parameters, potentially leading to incorrect inferences. To overcome these limitations, recent research has directly modeled hyper-relational facts. Some translation-based approaches~\cite{DBLP:conf/ijcai/WenLMCZ16,DBLP:conf/www/ZhangLMM18} define a hyper-relational fact through an attribute-value mapping. Meanwhile, tensor-based approaches~\cite{DBLP:journals/corr/abs-1906-00137,DBLP:conf/www/LiuY20} represent the truth space of hyper-relational facts using high-order tensors. More recently, neural network-based approaches~\cite{luo2023hahe,wang2023hyconve} have achieved significant performance improvements by leveraging neural networks to capture interactions between elements within hyper-relational facts.

\begin{figure}[t]
\centering  
\includegraphics[width=0.45\textwidth]{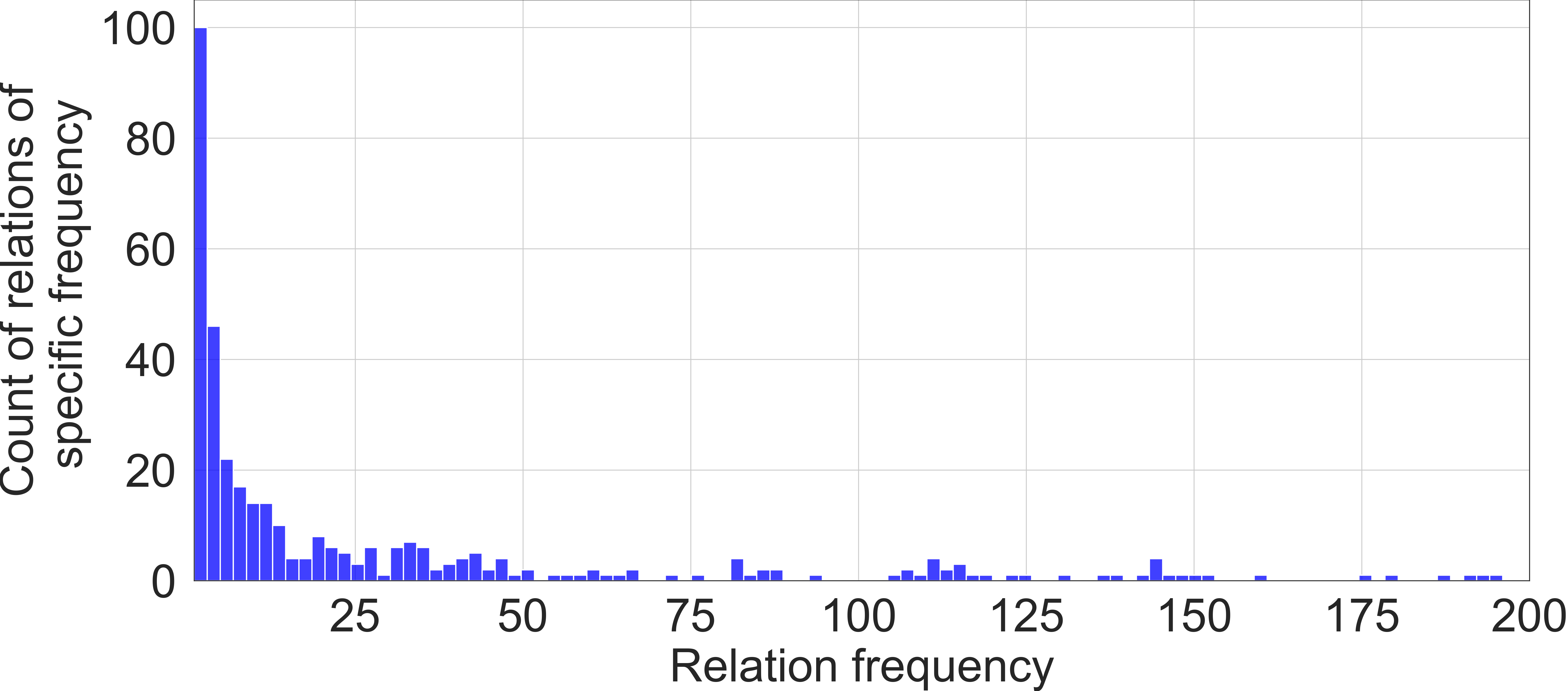}
\caption{The histogram of relation frequencies in WD50K.}
\label{Fig.main}
\end{figure}

However, current LPHFs methods often overlook the challenge of few-shot relations, even though these relations are prevalent in real-world KGs. For instance, in the benchmark dataset WD50K~\cite{DBLP:conf/www/RossoYC20}, it is observed that 32.5\% of relations have less than 5 instances (see Figure~\ref{Fig.main}). Moreover, real-world KGs are often dynamic, constantly introducing new relations with limited instances. While some existing studies focus on link prediction in few-shot scenarios~\cite{DBLP:conf/emnlp/ChenZZCC19,DBLP:conf/sigir/NiuLTGDLWSHS21}, they are designed for binary facts and cannot deal with attribute-value pairs, which are crucial for fully learning relation representations in hyper-relational facts. Therefore, there is an urgent need for methods that can effectively handle such scenarios.

Thus, we introduce a new task, called Few-Shot Link Prediction on Hyper-relational Facts (FSLPHFs). This task is to predict a missing entity in a hyper-relational fact associated with a relation $r$, given only a small number of support instances of $r$ (called support set). The main challenge of FSLPHFs lies in effectively learning the representation of $r$ in hyper-relational facts from these limited support instances. We tackle this challenge from two perspectives. Firstly, even though the few-shot relations have limited instances, the entities involved have background facts, which can be leveraged to generate few-shot relation representations. Secondly, taking inspiration from the success of meta-learning methods in the field of few-shot learning~\cite{metanetwork,DBLP:conf/icml/FinnAL17}, we can adjust relation representations using loss gradients of the support instances to obtain the essence knowledge of relations, which we refer to as meta relational information.

Based on these considerations, we design a model called MetaRH, which captures meta relational information from limited support instances to predict a missing entity in a query. MetaRH consists of three modules: relation learning, support-specific adjustment, and query inference. The relation learning module generates initial few-shot relation representations by aggregating entity background facts and encoding support instances. The support-specific adjustment module further adjusts relation representations based on the support set to obtain meta relational information. Finally, the query inference module predicts the missing entity in a query using the obtained meta relational information. Due to the lack of datasets designed for FSLPHFs, we construct three datasets based on existing LPHFs benchmark datasets. Through sufficient experiments, we demonstrate that MetaRH significantly outperforms existing models.

In summary, this paper makes the following contributions:
\begin{itemize}
  \item It propose a new task called Few-Shot Link Prediction on Hyper-relational Facts (FSLPHFs), which is practical in real-world scenarios.
  \item To tackle FSLPHFs, we propose the MetaRH method, which captures meta relational information from limited support instances to predict the missing entity in a query.
  \item Three datasets based on existing LPHFs benchmark datasets are constructed, providing valuable resources for evaluating FSLPHFs and further research in this area.
  \item Through extensive experiments conducted in various settings, we demonstrate that MetaRH achieves superior results in FSLPHFs, showcasing its effectiveness and potential for practical applications.
\end{itemize}

\section{Related Work}\label{sec:related_work}
Our work is the first to tackle few-shot link prediction on hyper-relational facts, filling a gap in the existing literature. The closest related research areas are Link Prediction on Hyper-relational Facts (LPHFs) and Few-Shot Link Prediction on Binary Facts (FSLPBFs). 
\subsection{Related Work on LPHFs}
Existing LPHFs works can be categorized into three groups: translation-based, tensor-based, and neural network-based. 

\textbf{Translation-based models} embed entities and relations into a low-dimensional space and make predictions by translating entities through relations. In the initial work, m-TransH~\cite{DBLP:conf/ijcai/WenLMCZ16} represents a hyper-relational fact as a mapping from a sequence of attributes to their corresponding values and models it to obtain the truth value of the fact. RAE~\cite{DBLP:conf/www/ZhangLMM18} enhances m-TransH by considering entity correlations.

\textbf{Tensor-based models} utilize a high-order tensor to represent the truth space of facts and predict new links by reconstructing the tensor. Due to their effectiveness on binary facts, researchers have extended them to handle hyper-relational facts. Some examples of such extensions include m-DistMult~\cite{DBLP:journals/corr/abs-1906-00137}, HypE~\cite{DBLP:journals/corr/abs-1906-00137}, and GETD~\cite{DBLP:conf/www/LiuY20}, which are generalizations of DistMult~\cite{DBLP:journals/corr/abs-1906-00137}, SimplE~\cite{DBLP:conf/nips/Kazemi018}, and TuckER~\cite{DBLP:conf/emnlp/BalazevicAH19}, respectively.

\textbf{Neural network-based models} utilize neural networks to capture element interactions in hyper-relational facts. NaLP~\cite{DBLP:conf/www/GuanJWC19} represents hyper-relational facts as attribute-value pairs and models their correlation using a fully connected neural network. HINGE~\cite{DBLP:conf/www/RossoYC20} and NeuInfer~\cite{DBLP:conf/acl/GuanJGWC20} represent hyper-relational facts as a primary triple with auxiliary attribute-value pairs and evaluate the validity and compatibility of these two components. StarE~\cite{DBLP:conf/emnlp/GalkinTMUL20} proposes a graph representation learning mechanism for hyper-relational facts, enhancing the communication from the auxiliary attribute-value pairs to the primary triple. GRAN~\cite{DBLP:conf/acl/WangWLZ21} extends StarE to represent hyper-relational facts as heterogeneous graphs and uses edge-biased attention layers to encode these graphs. Since StarE and GRAN only consider global or local structures in KGs, HAHE~\cite{luo2023hahe} further proposes a hierarchical attention mechanism that includes global and local attention. ShrinkE~\cite{xiong2023shrinking} extend Box~\cite{abboud2020boxe} to capture essential inference patterns of hyper-relational facts. The above models generally encode facts in Euclidean space, making it challenging to preserve the hierarchical relationships of entities. PolygonE~\cite{DBLP:conf/aaai/YanSY2022} embeds hyper-relational facts as gyro-polygons in hyperbolic poincaré ball and designs a vertex-gyrocentoid optimization goal to measure fact validity. Additionally, HyConvE~\cite{wang2023hyconve} exploits the powerful learning ability of convolutional neural networks for LPHFs.

\subsection{Related Work on FSLPBFs}
Existing FSLPBFs works can be categorized into two groups: metric learning-based and meta learning-based.

\textbf{Metric learning-based models} match queries to support instances and make predictions based on the match values. GMatching~\cite{DBLP:conf/emnlp/XiongYCGW18} enhances entity representations and learns a matching processor for prediction. FSRL~\cite{DBLP:conf/aaai/ZhangYHJLC20} extends GMatching by integrating information from multiple instances rather than relying on just one. FAAN~\cite{DBLP:conf/emnlp/ShengGCYWLX20} further introduces an adaptive attention mechanism that selectively focuses on entity properties.

\textbf{Meta learning-based models} calculate the gradient on support instances and quickly optimize parameters. MetaR~\cite{DBLP:conf/emnlp/ChenZZCC19} transfers relation information from support instances to queries using relation gradients. MetaP~\cite{DBLP:conf/sigir/JiangGL21} utilizes more efficient convolutional filters and proposes a validity balance mechanism of negative samples. GANA~\cite{DBLP:conf/sigir/NiuLTGDLWSHS21} combines MAML~\cite{DBLP:conf/icml/FinnAL17} and TransH~\cite{DBLP:conf/aaai/WangZFC14} to predict few-shot complex relations.

Existing LPHFs models overlook few-shot relations, while FSLPBFs models focus on binary facts and cannot handle hyper-relational facts.

\section{Problem Formulation}
In this section, we provide the definitions of hyper-relational facts, link prediction on hyper-relational facts, and few-shot link prediction on hyper-relational facts in turn.
\newtheorem{myDef}{\textbf{Definition}}
\begin{myDef}
{ \textbf{Hyper-relational facts}}{ are composed of a primary triple $(h, r, t)$ and several auxiliary attribute-value pairs $\{(a_i, v_i)\}^m_{i=1}$~\cite{DBLP:conf/www/RossoYC20}. Here, $r, a_i, ..., a_m\in R$ and $h, t, v_1, ..., v_m \in E$, with $m$ denoting the number of auxiliary attribute-value pairs, $E$ representing the set of entities and values, and $R$ denoting the set of relations and attributes.}
\end{myDef}

For instance, the hyper-relational fact, \emph{Einstein} studied for a \emph{Doctorate's Degree} of \emph{Physics} at \emph{University of Zurich} from \emph{1901} to \emph{1905}, can be represented as:

\indent((Einstein, studied for, Doctorate's Degree),\{\\
\indent\indent\indent  (major, Physics),\\
\indent\indent\indent  (university, the University of Zurich),\\
\indent\indent\indent  (begin-time, 1901), (end-time, 1905)\}).
\begin{myDef}
{ \textbf{Link Prediction on Hyper-relational Facts (LPHFs)}} aims to predict one missing element in a hyper-relational fact~\cite{DBLP:conf/ijcai/WenLMCZ16}, such as predicting the tail entity of the incomplete hyper-relational fact $((h,r,?),\{(a_i, v_i)\}^m_{i=1})$. 
\end{myDef}
\begin{myDef}
{\textbf{Few-Shot Link Prediction on Hyper-relational Facts (FSLPHFs)}} aims to predict a missing entity\footnote{In this paper, we conduct FSLPHFs through tail entity prediction following~\cite{DBLP:conf/emnlp/XiongYCGW18,DBLP:conf/emnlp/GalkinTMUL20}, as head entity prediction can be transformed into tail entity prediction easily through inverse relations.} in a query in the query set $\mathcal{Q}_r=\{((h_q, r, ?), \{(a_{q_i}, v_{q_i})\}^m_{i=1})\}$ of a few-shot relation $r$, with $k$ support instances $\mathcal{S}_r=\{((h_s^j, r, t_s^j), \{(a_{s_{i}}^j, v_{s_{i}}^j)\}^m_{i=1})\}_{j=1}^k$ (referred to as the support set) given, called $k$-shot link prediction on hyper-relational facts.

\end{myDef}

The training process of FSLPHFs is based on a set of tasks, wherein each task is associated with a few-shot relation and has its support set and query set. Besides, each task has an entity candidate set, which contains candidate entities that satisfy possible entity types, following~\cite{DBLP:conf/emnlp/XiongYCGW18}. The testing process is performed on a set of new tasks, wherein each task is associated with a few-shot relation that has not appeared in the training process and has its support set and query sets.

Finally, we assume that the method has access to background data $\mathcal{B}$, which contains background facts about entities in $\mathcal{S}_r$, following ~\cite{DBLP:conf/emnlp/XiongYCGW18}. The background facts of an entity $e$ is a set of facts with $e$ as the head entity. To utilize $\mathcal{B}$ fully, inverse facts $\{((t, r^{-1}, h), \{(a_i, v_i)\}^m_{i=1})\}$ are added to $\mathcal{B}$. For example, for a fact in the background data, ((Game of Thrones (Q23572), cast member (P161), Ciarán Hinds (Q314892)), \{(character role (P453), Mance Rayder (Q5991029))\}), we add its inverse facts ((Ciarán Hinds (Q314892), cast member (P161)$^{-1}$, Game of Thrones (Q23572)), \{(character role (P453), Mance Rayder (Q5991029))\}) to the background data. To avoid data leakage, no few-shot relations exist in $\mathcal{B}$. 

\begin{figure*}
  \centering  
  \includegraphics[width=1\textwidth]{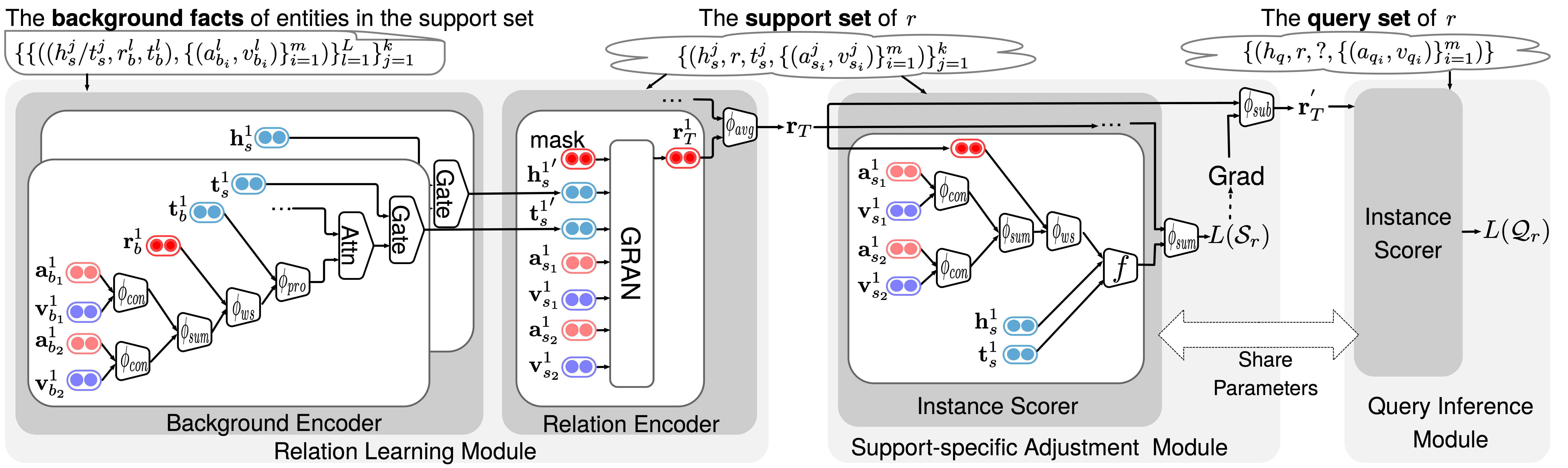}
  \caption{The overview of MetaRH model. To distinguish the different elements in hyper-relational facts, we use red for relations, blue for head and tail entities, pink for attributes, and purple for values.}
  \label{Fig.FSNR}
\end{figure*}

\section{The Proposed MetaRH Model}
We propose MetaRH to tackle FSLPHFs.  MetaRH consists of three modules: relation learning, support-specific adjustment, and query inference, as illustrated in Figure~\ref{Fig.FSNR}. To be clear, in Figure~\ref{Fig.FSNR} and what follows, we illustrate MetaRH with the background facts = $\{\{((h_s^j/t_s^j, r_b^l, t_b^l), \{(a_{b_{i}}^l, v_{b_{i}}^l)\}^m_{i=1})\}_{l=1}^L\}_{j=1}^k$, support set $\mathcal{S}_r$ = $\{((h_s^j, r, t_s^j), \{(a_{s_{i}}^j, v_{s_{i}}^j)\}^m_{i=1})\}_{j=1}^k$, and query set $\mathcal{Q}_r$ = $\{((h_q, r, ?), \{(a_{q_{i}}, v_{q_{i}})\}^m_{i=1})\}$. 

\subsection{Relation Learning Module}\label{relation_learning_module}
It is designed to obtain an initial representation of few-shot relation $r$ using a background encoder and a relation encoder.

\textbf{The background encoder} utilizes background facts to generate semantic-rich entity representations in the support set. To leverage priori knowledge from the background facts and enhance entity representations, we employ a Graph Neural Network with attention and gating mechanisms. 

For example, the semantic-rich representation ${\mathbf{t}_s^j}^{\prime}$ of entity $t_s^j$ is obtained by combining the initial entity representation ${\mathbf{t}_s^j}$ with its background fact representation $\mathbf{b}^l$, using attention values $\alpha$ and a gate value $g$, as follows:
\begin{equation}
    {\mathbf{t}_s^j}^{\prime}=\sigma\left(\sum_{l=1}^L{g \alpha^l \mathbf{b}^l}+\left(1-g\right) \mathbf{t}_s^j\right),   \label{enhanced_entity}
\end{equation}
\noindent where $L$ is the number of background facts per entity; the background facts of $\mathbf{t}_s^j$ is a set of facts in $\mathcal{B}$ with $\mathbf{t}_s^j$ as the head entity; $\alpha^l$ represents the attention value of background fact $b^l$; $\sigma$ is an activation function. The semantic-rich head entity representation ${\mathbf{h}_s^j}^{\prime}$ is obtained in the same manner as ${\mathbf{t}_s^j}^{\prime}$. Next, we provide a detailed explanation of the background fact representation $\mathbf{b}^l$, attention value $\alpha^l$, and gate value $g$.

The background fact representation $\mathbf{b}^l$ is calculated by aggregating all the elements in $b^l$. This involves using a weighted sum operation $\phi_{ws}$ and a project operation $\phi_{pro_2}$ to fuse the auxiliary attribute-value pairs representation $\mathbf{q}_{b}^l$ to the relation representation $\mathbf{r}_{b}^l$. The new relation representation is then combined with the tail entity representation $\mathbf{t}_{b}^l$ using a project operation $\phi_{pro}$ and a concatenate operation $\phi_{con}$. As a result, $\mathbf{b}^l$ is obtained as follows:
\begin{equation}
  \begin{split}
    \mathbf{b}^l & = \phi_{pro}(\phi_{con}(\phi_{ws}(\mathbf{r}_b^l, \phi_{pro_2}(\mathbf{q}_b^l)), \mathbf{t}_b^l))\\
    & = \mathbf{W}_{1}\left[(\tau \odot \mathbf{r}_b^l + \left(1-\tau\right) \odot \mathbf{W}_2 \mathbf{q}_b^l); \mathbf{t}_b^l\right] + \textbf{b}_1,
  \end{split}
  \label{con:c}
\end{equation}
\noindent where $b^l$ contains $((t_s^j, r_b^l, t_b^l), $ $\{(a_{b_{i}}^l, v_{b_{i}}^l)\}^m_{i=1})$; $\mathbf{W}_{1}$ and $\mathbf{W}_{2}$ are parameterized projection matrixes; $\textbf{b}_1$ is a parameterized bias; $\tau$ is a relation weight hyper-parameter; $\odot$ is a scalar product operation; inspired by Galkin et al.~\cite{DBLP:conf/emnlp/GalkinTMUL20}, $\mathbf{q}_{b}^l$ is obtained using a position invariant summation function $\phi_{sum}$ and a rotate function $\phi_{rot}$~\cite{sunrotate}, as follows:
\begin{equation}
  \begin{split}
    \mathbf{q}_b^l & = \phi_{sum}(\{\phi_{rot}(\mathbf{a}_{b_{i}}^l, \mathbf{v}_{b_{i}}^l)\}_{i=1}^m)\\
& = \sum_{(\mathbf{a}_{b_{i}}^l, \mathbf{v}_{b_{i}}^l) \in \{(\mathbf{a}_{b_{i}}^l, \mathbf{v}_{b_{i}}^l)\}^m_{i=1}} \phi_{rot}(\mathbf{a}_{b_{i}}^l, \mathbf{v}_{b_{i}}^l),  \label{con:q}
  \end{split}
\end{equation}
\noindent where $\mathbf{a}_{b_{i}}^l$ and $\mathbf{v}_{b_{i}}^l$ are the corresponding representations of $a_{b_{i}}^l$ and $v_{b_{i}}^l$, respectively.

To capture the most valuable information from background facts and filter out noisy facts, attention mechanisms and gating mechanisms are employed, referring to~\cite{DBLP:conf/emnlp/ShengGCYWLX20, DBLP:conf/sigir/NiuLTGDLWSHS21}. The attention value $\alpha^l$ of background fact $b^l$ is calculated by applying the softmax function on all absolute attention values, as follows:
\begin{equation}
\alpha^l=\frac{\exp \left(d^l\right)}{\sum_{l=1}^L \exp \left(d^l\right)}.
\end{equation}
\begin{equation}
d^l=\operatorname{LeakyReLU}\left(\textbf{U}^T_1\mathbf{b}^l\right), \label{con:d}
\end{equation}
\noindent where $d^l$ is the absolute attention value of $b^l$; $\operatorname{LeakyReLU}(\cdot)$~\cite{maas2013rectifier} is an activation function; $\textbf{U}_1$ is a weight vector.

To further filter out noisy background facts, gating mechanisms are implemented. The gate value $g$ of all background facts is calculated using a sigmoid function as follows:
\begin{equation}
g=\operatorname{sigmoid}\left(\textbf{U}_2^T \sum_{l=1}^L \alpha^l \mathbf{b}^l+b_{g}\right), \label{con:g}
\end{equation}
\noindent where $\textbf{U}_2$ is a weight vector, and $b_g$ is a scalar bias.

\textbf{The relation encoder} aims to generate few-shot relation representations with semantic-rich entity representations as input.

Many existing LPHFs models can generate relation representations based on hyper-relational facts. In this work, we select GRAN~\cite{DBLP:conf/acl/WangWLZ21} as the relation encoder, since it has shown effectiveness in the LPHFs task. Taking the support instance ((${{h_s^{j}}}$, $r$, ${{t_s^{j}}}$), $\{(a_{s_{i}}^j, v_{s_{i}}^j)\}^m_{i=1}$) as an example, the semantic-rich instance representation, ((${{\mathbf{h}_s^{j}}^\prime}$, $mask$, ${{\mathbf{t}_s^{j}}^\prime}$), $\{(\mathbf{a}_{s_{i}}^j, \mathbf{v}_{s_{i}}^j)\}^m_{i=1}$), is represented as a heterogeneous graph $\mathcal{G}$, where $mask$ is a special token denoting the few-shot relation $r$. More details on GRAN can be found in~\cite{DBLP:conf/acl/WangWLZ21}. Then graph $\mathcal{G}$ is then processed through a stack of $D$ GRAN blocks, as follows:
\begin{equation}
  \mathcal{G}^{d}=\operatorname{GRAN}\left(\mathcal{G}^{d-1}\right), d=1,2, \cdots, D, 
  \label{con:gran}
\end{equation}
\noindent where $\mathcal{G}^{d}$ is the hidden state after $d$-th layer. The representation of $mask$ in the last layer is selected as the few-shot relation representation $\mathbf{r}_{T_j}$.

The relation representations of other support instances are obtained similarly. The few-shot relation representation $\mathbf{r}_{T}$ of the current task is obtained through an average operation $\phi_{avg}$ :
\begin{equation}
  \mathbf{r}_{T} = \phi_{avg}(\{\mathbf{r}_{T_j}\}_{j=1}^k) = \frac{1}{k}\sum_{j=1}^k{\mathbf{r}_{T_j}},    \label{initial_r}
\end{equation}
\noindent where $k$ is the number of support instances.
\subsection{Support-specific Adjustment Module}\label{task_updating_module}
The previous module generates a few-shot relation representation. However, it is coarse due to the simple aggregation operation in Equation~\ref{initial_r}. Thus, the support-specific adjustment module is designed to obtain meta relational information that represents common knowledge within the task. This module utilizes the gradient on support instances to guide the adjustment of the coarse relation representation based on an instance scorer. Before introducing the adjustment of the relation representation, we first introduce the instance scorer and the loss of the support set.

\begin{algorithm}[tb]
  \caption{The training process of MetaRH.}
  \textbf{Input}: Training tasks $\mathcal{T}_{training}$; Initial parameters.
  \begin{algorithmic}[1] 
      \REPEAT
      \STATE Sample mini-batch tasks $\mathcal{F}_{t}$ from $\mathcal{T}_{training}$.\\
      \FOR{each task in $\mathcal{F}_t$}
      \STATE Sample few-shot instances as $S_r$.\\
      \STATE Sample a batch of instances as $Q_r$.\\
      \STATE Get the background facts of $S_r$.\\
      \STATE Generate semantic-rich entity representations in $S_r$ (Equation~\ref{enhanced_entity}$\sim$Equation~\ref{con:g}).\\
      \STATE Generate an initial few-shot relation representation $\mathbf{r}_T$ (Equation~\ref{con:gran}$\sim$Equation~\ref{initial_r}).\\
      \STATE Calculate the loss of $\mathcal{S}_r$ (Equation~\ref{con:q_s}$\sim$Equation~\ref{support_loss}).\\
      \STATE Generate meta relational information $\mathbf{r}_T^\prime$ (Equation~\ref{up_gard}).\\
      \STATE Calculate the loss of $\mathcal{Q}_r$ (Equation~\ref{query_loss}).\\
      \ENDFOR
      \STATE Update model parameters.\\
      \UNTIL{process completes maximum times.}
  \end{algorithmic}\label{algorithm1}
\end{algorithm}

\textbf{The instance scorer} evaluates the semantic connections between few-shot relations and other elements in instances. Previous research in FSLPBFs~\cite{DBLP:conf/emnlp/ChenZZCC19} has shown that the translation-based model TransE~\cite{NIPS2013_1cecc7a7} performs well as an instance scorer. Therefore, we adopt a translation-based instance scorer in this work, which is designed as follows:

Taking the support instance $((h_s^j, r, t_s^j), $ $ \{(a_{s_{i}}^j, v_{s_{i}}^j)\}^m_{i=1})$ as an example, we first calculate the auxiliary attribute-value pairs representation $\mathbf{q}_s$ similarly as $\mathbf{q}_b^l$ (see Equation~\ref{con:q}) and aggregate it to $\mathbf{r}_T$, as follows:
\begin{equation}
  \begin{split}
    \mathbf{q}_s^l & = \phi_{sum}(\{\phi_{rot}(\mathbf{a}_{s_{i}}^l, \mathbf{v}_{s_{i}}^l)\}_{i=1}^m)\\
    & = \sum_{(\mathbf{a}_{s_{i}}^l, \mathbf{v}_{s_{i}}^l) \in \{(\mathbf{a}_{s_{i}}^l, \mathbf{v}_{s_{i}}^l)\}^m_{i=1}} \phi_{rot}(\mathbf{a}_{s_{i}}^l, \mathbf{v}_{s_{i}}^l),  \label{con:q_s}
  \end{split}
\end{equation}
\begin{equation}
  \mathbf{r}_s^j=\phi_{ws}(\mathbf{r}_T, \phi_{pro_2}(\mathbf{q}_s^j))=\tau \odot \mathbf{r}_T + (1-\tau) \odot \mathbf{W}_2 \mathbf{q}_s^j,  \label{support_relation}
\end{equation}
\noindent where $\mathbf{a}_{s_{i}}^j$, $\mathbf{v}_{s_{i}}^j$ are the corresponding representations of $a_{s_{i}}^j$, $v_{s_{i}}^j$ respectively; $\mathbf{r}_s^j$ is the new relation representation that incorporates the representation of auxiliary attribute-value pairs.

Then, following TransE, the score of the support instance is calculated as follows:
\begin{equation}
  f{ \left( \mathbf{h}_s^j, \mathbf{r}_s^j, \mathbf{t}_s^j \right) }=\left\|\mathbf{h}_s^j+ \mathbf{r}_s^j - \mathbf{t}_s^j\right\|, \label{support_score}
\end{equation}
\noindent where $\mathbf{h}_s^{j}$, $\mathbf{t}_s^{j}$ are the corresponding representations of $h_s^{j}$, $t_s^{j}$ respectively; $||\mathbf{x}||$ is the L2 norm of vector $\mathbf{x}$; this function is denoted as ``$f$'' in Figure~\ref{Fig.FSNR}. To obtain accurate common knowledge from the support set, the initial entity representations are used instead of rich-semantic ones.

\textbf{The loss of support set} is defined as follows:
\begin{equation}
L\left(\mathcal{S}_{r}\right)\!=\!\!\!\!\!\sum_{\left(\mathbf{h}_s^j, \mathbf{r}_s^j, \mathbf{t}_s^j \right) \in \mathcal{S}_{r}}\!\!\!\!\left[\mu\!\!+\!\!f{\left(\mathbf{h}_s^j, \mathbf{r}_s^j, \mathbf{t}_s^j \right)}\!\!-\!\!f{\left(\mathbf{h}_s^j, \mathbf{r}_s^j, \mathbf{t}_s^{j^{\prime\prime}} \right)}\right]_{+},    \label{support_loss}
\end{equation}
\noindent where $[x]_{+} = max[0, x]$ is hinge loss; $\mu$ is a margin hyper-parameter; $\mathbf{t}_s^{j^{\prime\prime}}$ is generated by randomly corrupting tail entities of support instances.

\textbf{The adjustment of $\mathbf{r}_{T}$} is guided by the gradient of $\mathbf{r}_{T}$, which indicates how $\mathbf{r}_{T}$ should be adjusted, as $L(S_r)$ represents the ability of the instance scorer to encode the truth values of instances. Therefore, we obtain meta relational information $\mathbf{r}_{T}^{\prime}$, as follows:
\begin{equation}
\begin{split}
  \mathbf{r}_{T}^{\prime} & = \phi_{sub}(\mathbf{r}_{T}, Grad(\mathbf{r}_T))\\
  & = \mathbf{r}_{T}-\beta \frac{\mathrm{d} L\left(\mathcal{S}_{r}\right)}{\mathrm{d} \mathbf{r}_{T}},  \label{up_gard}
\end{split}
\end{equation}
\noindent where $\phi_{sub}$ is a subtraction operation; $Grad(\mathbf{r}_T)$ is the gradient of $\mathbf{r}_{T}$; $\beta$ indicates the step size of the gradient when adjusting $\mathbf{r}_{T}$. 

\subsection{Query Inference Module}\label{query_inference_module}
The query inference module predicts the missing entity in a query using an instance scorer. To enhance the training efficiency of MetaRH, the instance scorer in the query inference module adapts the same structure and shares parameters as the instance scorer introduced in the support-specific adjustment module (see Section \ref{task_updating_module}).

\textbf{The loss of query set} is computed similarly to $L(\mathcal{S}_r)$ (see Equation~\ref{support_loss}), as follows:
\begin{equation}
  L (\mathcal{Q}_{r} )\!=\!\!\!\!\sum_{ \left(\mathbf{h}_q, \mathbf{r}_q, \mathbf{t}_q \right) \in \mathcal{Q}_{r}}\!\!\!\! \left[\mu\!\!+\!\!f{ \left(\mathbf{h}_q, \mathbf{r}_q, \mathbf{t}_q \right)}\!\!-\!\!f{ \left(\mathbf{h}_q, \mathbf{r}_q, \mathbf{t}_q^{{\prime\prime}} \right)}  \right]_{+},   \label{query_loss}
\end{equation}
\noindent where $\mathbf{h}_q$, $\mathbf{t}_q$ is the corresponding representations of $h_q$, $t_q$ respectively; the new relation representation $\mathbf{r}_q$ is obtained by combining the representation of attribute-value pairs and meta relational information, similar to $\mathbf{r}_s^j$ (see Equation~\ref{support_relation}); $\mathbf{t}_q^{{\prime\prime}}$ is the negative entity representation, which is generated in a similar way as $\mathbf{t}_s^{j^{\prime\prime}}$ (see Equation~\ref{support_loss}). For further details on the training process, refer to Algorithm~\ref{algorithm1}.

\begin{table*}\scriptsize
  \centering
  \begin{tabular}{llllllllll}
  \hline 
  \text { Dataset } & \text { \#E } & \text { \#R } & \text { \#E-q } & \text { \#R-q } & \text { \#B-facts } & \text { B-N-rate } & \text { \#F-facts } & \text { F-N-rate } & \text { \#Tasks } \\
  \hline 
  \text { F-WikiPeople } & 40529 & 237 & 4663 & 75 & 314670 & 9.1\% & 4470 & 1.5\% & 30 \\
  \text { F-JF17K } & 19721 & 480 & 4928 & 127 & 86415 & 49.3\% & 5157 & 19.2\% & 52 \\
  \text { F-WD50K } & 43802 & 697 & 10242 & 85 & 358439 & 13.8\% & 21214 & 1.8\% & 118 \\
  \hline
  \end{tabular}
  \caption{Statistics of the constructed datasets.}
  \label{Tab.datasets}
\end{table*}
\section{Experiments}

\subsection{Datasets}
Since there is no dataset specifically designed for FSLPHFs, we construct three new datasets, F-WikiPeople, F-JF17K, and F-WD50K, by modifying existing LPHFs benchmark datasets WikiPeople~\cite{DBLP:conf/cikm/GuanJWC18}, JF17K~\cite{DBLP:conf/ijcai/WenLMCZ16}, and WD50K~\cite{DBLP:conf/www/RossoYC20}, respectively. These LPHFs datasets are derived from real-world KGs and are widely used in LPHFs. Specifically, the JF17K dataset is derived from Freebase~\cite{bollacker2008freebase}, and the Wikipeople and WD50K datasets is derived from Wikidata~\cite{vrandevcic2014wikidata}. The Wikipeople dataset stores a large number of facts related to people, while WD50K stores a large number of facts in which head entities appear in the well-known knowledge graph FB15K-237~\cite{NIPS2013_1cecc7a7}. We believe that our proposed datasets will provide valuable resources for further research in this field.

New FSLPHFs datasets are constructed as follows:
\begin{itemize}
\item Select relations with 20-1000 instances\footnote{The lower boundary is to have enough facts for evaluation. The upper boundary is to retain some facts to be used as background data.} as few-shot relations from each existing dataset.
\item Get few-shot data by retrieving the instances of few-shot relations.
\item Remove instances with few-shot relations in auxiliary attribute-value pairs from the few-shot data to prevent data leakage.
\item Get background data $\mathcal{B}$ by retrieving the instances of entities in the few-shot data from the original dataset.
\item Remove instances containing few-shot relations from $\mathcal{B}$ to prevent data leakage.
\item Divide the few-shot data into training tasks $\mathcal{T}_{training}$ , validation tasks $\mathcal{T}_{validation}$, and testing tasks $\mathcal{T}_{testing}$, in the proportion of 85\%: 5\%: 10\%, following~\cite{DBLP:conf/emnlp/XiongYCGW18}.
\end{itemize}

Table~\ref{Tab.datasets} provides statistics of the constructed datasets, including counts for various elements: \#X is the number of X, E-q and R-q denote the values and attributes in auxiliary attribute-value pairs respectively, B-facts and F-facts denote the facts in background data and few-shot data respectively, B-N-rate and F-N-rate denote the proportion of hyper-relational facts in background data and few-shot data respectively, and Tasks denotes the few-shot tasks. 

\begin{table*}[h]\scriptsize
  \centering
  \begin{tabular}{c|cccc|cccc|cccc}
  \hline 
  \multirow{2}{*}{Method} & \multicolumn{4}{c|}{\text { F-WikiPeople }} & \multicolumn{4}{c|}{\text { F-JF17K }} & \multicolumn{4}{c}{\text { F-WD50K }} \\
  \cline { 2 - 13 }
   & MRR & Hits@10 & Hits@5  & Hits@1 & MRR & Hits@10 & Hits@5  & Hits@1 & MRR & Hits@10 & Hits@5 & Hits@1 \\
  \hline 
  \text { m-TransH } & 0.197 & 0.403 & 0.309 & 0.101 & 0.045 & 0.076 & 0.047 & 0.021 & 0.051 & 0.081 & 0.047 & 0.015\\
  \text { HypE } & 0.291 & 0.516 & 0.368 & 0.195 & 0.042 & 0.111 & 0.040 & 0.014 & 0.051 & 0.106 & 0.063 & 0.024 \\
  \text { PolygonE } & 0.231 & 0.388 & 0.268 & 0.137 & 0.057 & 0.192 & 0.134 & 0.025 & 0.052 & 0.107 & 0.060 & 0.018\\
 \text { NeuInfer } & 0.289 & 0.581 & 0.455 & 0.155 & 0.092 & 0.156 & 0.111 & 0.061 & 0.133 & 0.231 & 0.180 & 0.078\\
 \text { HINGE } & 0.333 & 0.439 & 0.276 & 0.277 & 0.084 & 0.124 & 0.095 & 0.064 & 0.154 & 0.267 & 0.213 & 0.089 \\
 \text { StarE } & 0.286 & 0.558 & 0.471 & 0.118 & 0.117 & 0.151 & 0.135 & 0.090 & 0.102 & 0.177 & 0.134 & 0.057 \\
 \text { GRAN } & 0.287 & 0.432 & 0.374 & 0.209 & 0.119 & 0.157 & 0.124 & 0.101 & 0.126 & 0.222 & 0.162 & 0.077 \\
 \text { ShrinkE } & 0.314 & 0.504 & 0.421 & 0.221 & 0.051 & 0.123 & 0.063 & 0.020 & 0.046 & 0.081 & 0.059 & 0.024 \\
 \text { HyConvE } & 0.364 & \underline{0.621} & 0.428 & 0.272 & 0.177 & 0.289 & 0.234 & \underline{0.123} & 0.086 & 0.176 & 0.118 & 0.038 \\
 \text { HAHE } & \underline{0.392} & 0.583 & 0.480 & \underline{0.306} & \underline{0.182} & \underline{0.293} & \underline{0.276} & 0.117 & 0.157 & 0.265 & 0.206 & 0.102 \\
  \hline 
 \text { FAAN } & 0.266 & 0.550 & 0.412 & 0.092 & 0.032 & 0.090 & 0.034 & 0.003 & 0.116 & 0.226 & 0.166 & 0.059 \\
 \text { MetaR } & 0.282 & 0.556 & 0.459 & 0.147 & 0.047 & 0.086 & 0.055 & 0.022 & 0.108 & 0.183 & 0.139 & 0.064 \\
 \text { GANA } & 0.341 & 0.475 & 0.371 & 0.275 & 0.074 & 0.218 & 0.130 & 0.016 & \underline{0.176} & 0.313 & 0.246 & 0.100 \\
 \hline
 \text { ChatGPT } & - & 0.584 & \textbf{0.548} & \textbf{0.358} & - & 0.165 & 0.140 & 0.093 & - & \textbf{0.548} & \textbf{0.474} & \textbf{0.237} \\
 \text { MetaRH } & \textbf{0.415} & \textbf{0.644} & \underline{0.500} & \underline{0.318} & \textbf{0.214} & \textbf{0.329} & \textbf{0.292} & \textbf{0.141} & \textbf{0.192} & \underline{0.340} & \underline{0.278} & \underline{0.109} \\
  \hline
  \end{tabular}
  \caption{Few-shot link prediction performance on hyper-relational facts.}
  \label{N-ary-epx}
\end{table*}

\subsection{Experimental Settings}\label{implementation_details}
\textbf{Baselines}. Due to the lack of models designed specifically for FSLPHFs, MetaRH is primarily compared with LPHFs and FSLPBFs models: (1) representative or state-of-the-art LPHFs models: m-TransH~\cite{DBLP:conf/ijcai/WenLMCZ16}, HypE~\cite{DBLP:journals/corr/abs-1906-00137}, NeuInfer~\cite{DBLP:conf/acl/GuanJGWC20}, HINGE~\cite{DBLP:conf/www/RossoYC20}, StarE~\cite{DBLP:conf/emnlp/GalkinTMUL20}, GRAN~\cite{DBLP:conf/acl/WangWLZ21}, PolygonE~\cite{DBLP:conf/aaai/YanSY2022}, ShrinkE~\cite{xiong2023shrinking}, HyConvE~\cite{wang2023hyconve}, and HAHE~\cite{luo2023hahe}; (2) advanced FSLPBFs models: MetaR~\cite{DBLP:conf/emnlp/ChenZZCC19}, FAAN~\cite{DBLP:conf/emnlp/ShengGCYWLX20}, and GANA~\cite{DBLP:conf/sigir/NiuLTGDLWSHS21}. More details on these models can be found in Section~\ref{sec:related_work}. Additionally, MetaRH is compared with ChatGPT~\footnote{https://openai.com/blog/chatgpt/}, a recently prominent Large Language Model (LLM). 

\textbf{Evaluation metrics} used are Hits@k and Mean Reciprocal Rank (MRR), with k = 1, 5, 10, following~\cite{DBLP:conf/emnlp/XiongYCGW18}. The hits@k metric is the proportion of the correct answer ranked within the top k, while the MRR metric is the average of the reciprocal rank of the correct answer. Higher values of MRR and Hits@k indicate better performance.

\textbf{Implementation details}. Hyper-parameters of MetaRH are selected within the following ranges: The embedding dimension $\in\{50, 100\}$, the batch size of tasks per epoch $\in\{128, 256, 512, 1024, 2048\}$, the batch size of queries per task $\in\{1,2,3,4,5\}$, the learning rate $\in\{5e-3,1e-3,5e-4,1e-4\}$, the maximum number of background facts per entity $\in \{10,20,30,50\}$, the margin $\mu \in\{1,2,3,4,5\}$, and the relation weight $\tau \in[0.0,1.0]$ with step is 0.1. The embedding layer is initialized with pre-trained embeddings trained on the background data with HINGE, following~\cite{DBLP:conf/emnlp/XiongYCGW18}. The Adam optimizer~\cite{kingma2015adam} is used to optimize the model. We conduct all experiments in the 5-shot scenario, that is, $k$ is set to 5. The Code and datasets of this paper can be found at https://github.com/JiyaoWei/MetaRH.

To ensure a fair comparison, LPHFs baselines are trained using all facts in \{$\mathcal{T}_{training}$, $\mathcal{S}_r \in$ $\mathcal{T}_{validation}\cup\mathcal{T}_{testing}$, $\mathcal{B}$\}, while FSLPBFs baselines are trained using the same $\mathcal{T}_{training}$ and $\mathcal{B}$ employed by MetaRH. The hyper-parameters of all baselines, except ChatGPT, are tuned on each experimental dataset.

\begin{figure}[t]
  \centering  
  \includegraphics[width=0.45\textwidth]{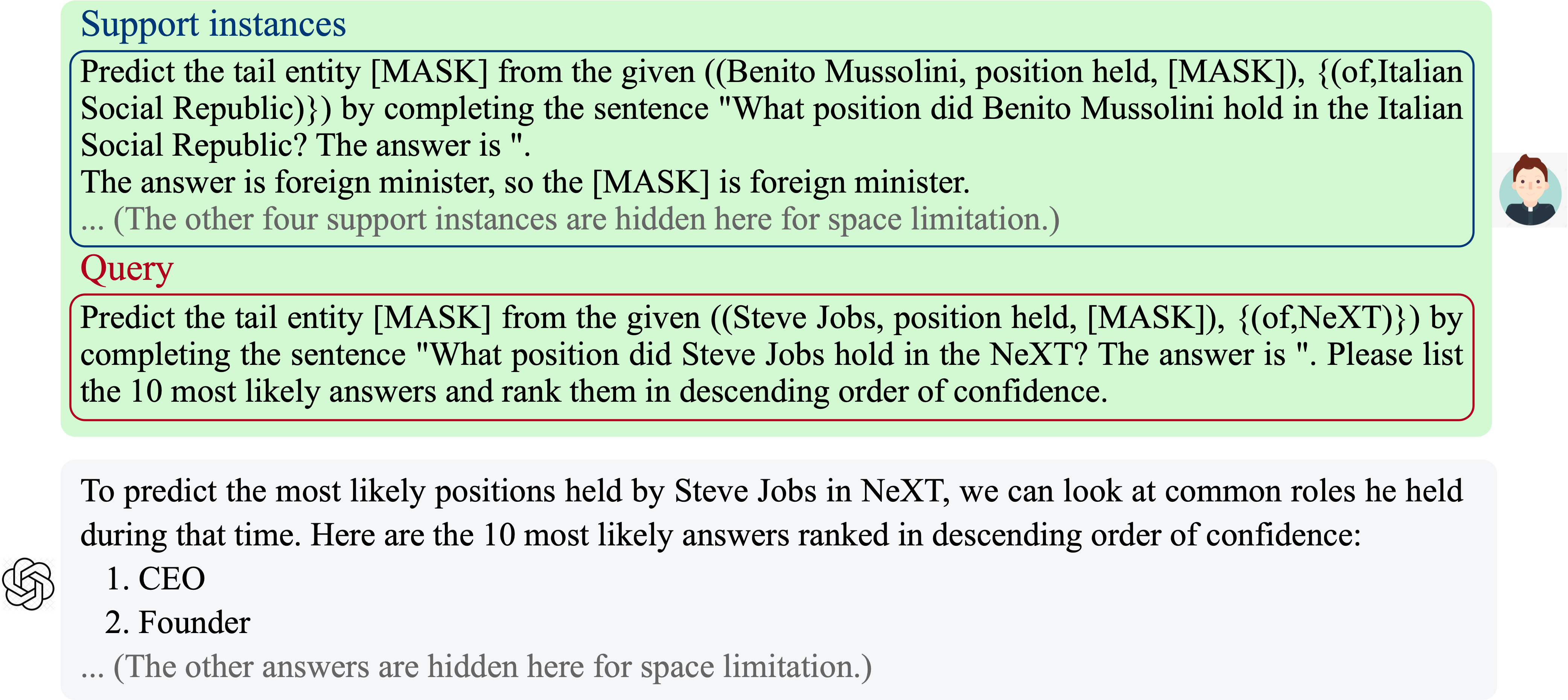}
  \caption{\!\!An example of ChatGPT prompts for FSLPHFs.}
  \label{Fig.ChatGPT2}
\end{figure}

For ChatGPT, to enhance the persuasiveness of the experiments, we manually constructed the prompt of ChatGPT for each query following Zhu et al.~\cite{zhu2023llms}. Furthermore, the prompt is modified to produce multiple candidates for a more in-depth comparative analysis. Specifically, we added ``Please list the 10 most likely answers and rank them in descending order of confidence.'' at the end of the current prompt. ChatGPT would generate 10 rows, each representing one candidate answer, i.e. ``1. $\{$candidate$\}$$\backslash$n 2. $\{$candidate$\}$$\backslash$n ... 10. $\{$candidate$\}$''. $\{$candidate$\}$ indicates a generated answer. For the metric Hits@10, if the true entity appears in any of the answers generated by ChatGPT, the prediction is considered correct. For the metric Hits@k, if the true entity appears in the first k generated answers, the prediction is considered correct. Figure~\ref{Fig.ChatGPT2} illustrates the format of the prompts, which include support instances and a query. Given that the accuracy of the responses generated by ChatGPT is the only metric available, we only show its Hits metrics.

\begin{table*}\scriptsize
  \setlength{\tabcolsep}{1mm}{
      \centering
      \begin{tabular}{l|llll|llll|llll}
      \hline
      ~ & \multicolumn{4}{c|}{\text{F-WikiPeople}}& \multicolumn{4}{c|}{\text{F-JF17K}}& \multicolumn{4}{c}{\text{F-WD50K}}\\
      \cline { 2 - 13 }
      \text{Methods} & \text{MRR} & \text{Hits@10} & \text{Hits@5} & \text{Hits@1}  & \text{MRR} & \text{Hits@10} & \text{Hits@5} & \text{Hits@1} & \text{MRR} & \text{Hits@10} & \text{Hits@5} & \text{Hits@1}\\
      \hline
      \text { MetaRH } & \textbf{0.415} & \textbf{0.644} & \textbf{0.500} & \textbf{0.318} & \textbf{0.214} & \textbf{0.329} & \textbf{0.292} & \textbf{0.141} & \textbf{0.192} & \textbf{0.340} & \textbf{0.278} & \textbf{0.109} \\
      \hline
      -background & 0.382 & 0.540 & 0.421 & 0.304 & 0.199 & 0.306 & 0.285 & 0.109 & 0.177 & 0.326 & 0.265 & 0.091 \\
      -adjustment & 0.328 & 0.450 & 0.414 & 0.261 & 0.182 & 0.299 & 0.270 & 0.104 & 0.152 & 0.285 & 0.219 & 0.082 \\
      \hline
      \end{tabular}
      \caption{Experimental results of the ablation study.}
      \label{ablation}

  }
\end{table*}

\subsection{Experimental Results and Analysis}
The experimental results for all three datasets are displayed in Table~\ref{N-ary-epx}, highlighting the best results in bold and the second-best results underlined. We have the following observations:

\textbf{Comparing MetaRH with LPHFs baselines}, MetaRH outperforms all existing models across all three datasets. For instance, in terms of the Hits@10 metric, MetaRH achieves improvements of 2.3\% on the F-WikiPeople dataset (3.7\% relative improvement ), 3.9\% on the F-JF17K dataset (13.3\% relative improvement), and 7.3\% on the F-WD50K dataset (27.3\% relative improvement). The remarkable success of MetaRH can be attributed to learning meta relational information. This targeted design empowers MetaRH to effectively capture the essential knowledge of relations with limited instances. Notably, MetaRH demonstrates the most significant performance improvement on the F-WD50K dataset, which involves the maximum number of training tasks. This observation suggests that the more training tasks there are, the stronger MetaRH's ability to learn meta relational information.

\textbf{Comparing MetaRH with FSLPBFs baselines}, MetaRH outperforms existing models across all three datasets. For instance, in terms of the Hits@10 metric, MetaRH achieves improvements of 8.8\% on F-Wikidata (15.8\% relative improvement), 11.1\% on the F-JF17K dataset (50.9\% relative improvement), and 2.7\% on the F-WD50K dataset (8.6\% relative improvement). This improvement demonstrates the effectiveness of leveraging auxiliary attribute-value pairs in few-shot relation learning. Moreover, MetaRH achieves a significant improvement on the F-JF17K dataset, which has a high proportion of hyper-relational facts, further emphasizing the importance of using auxiliary attribute-value pairs. 

\textbf{Comparing MetaRH with ChatGPT}, MetaRH performs better on the F-JF17K dataset but falls short on the F-WikiPeople and F-WD50K datasets, considering most metrics. This performance discrepancy may be due to the variance in data sources. The F-JF17K dataset is derived from Freebase, while the F-WikiPeople and F-WD50K datasets are derived from Wikidata. MetaRH achieves a relative improvement of up to 51.6\% in the metric hits@1 on the F-JF17K dataset, where 87\% of the knowledge is domain-specific knowledge, including film and sport. MetaRH does not perform well on the F-WikiPeople dataset and F-WD50K dataset, since these two datasets are derived from Wikidata, storing a large amount of generalized domain knowledge such as geography, country, etc. This indicates that knowledge graph models are still necessary in the real scenario currently. They achieve better performance on the reasoning task in non-generalized domains, as demonstrated by the LLM survey~\cite{pan2024unifying}. We also speculate that the unpublished training datasets used by ChatGPT include Wikidata or related datasets such as Wikipedia, but not Freebase. The opacity of the training data seriously affects its practical applications. Additionally, crafting high-quality prompts is crucial, but it is laborious and requires expert experience. For the metric MRR, calculating it of ChatGPT on link prediction is still a challenge since we can only check if the answer is in the response of ChatGPT but is almost impossible to get the rank for each answer.

\begin{figure}[t]
  \centering  
  \includegraphics[width=0.45\textwidth]{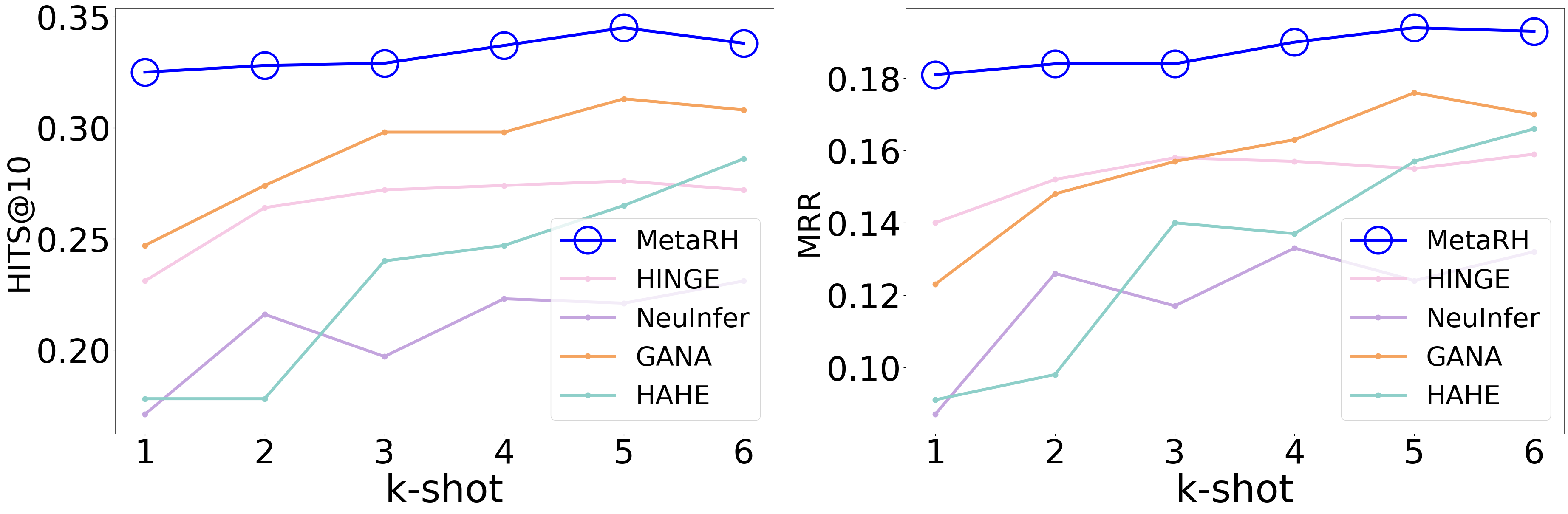}
  \caption{Impact of the few-shot size on the F-WD50K dataset.}
  \label{Fig.k-shot}
\end{figure}

Furthermore, we conduct experiments on the largest dataset F-WD50K to analyze the impact of the $k$-shot setting. We follow Sheng et al.~\cite{DBLP:conf/emnlp/ShengGCYWLX20} to vary $k$ from 1 to 6. We compare MetaRH with several competitive baseline models, namely HINGE, NeuInfer, GANA, and HAHE. The results, depicted in Figure~\ref{Fig.k-shot}, show that MetaRH consistently outperforms the baseline models across various $k$ values. Additionally, the performance of baselines does not plateau. We speculate that it is due to that they are less capable of data utilization. Specifically, the baselines for link prediction on hyper-relational facts (e.g., HINGE, Nueinfer, and HAHE) cannot effectively capture the essential knowledge of relations with limited instances. The few-shot link prediction baseline (e.g., MetaR) cannot leverage auxiliary attribute-value pairs. They all have data utilization issues and are sensitive to different values of $k$.

\subsection{Ablation Studies}
The two essential components of MetaRH are the background encoder and support-specific adjustment module. To evaluate their necessity, ablation studies are conducted on all three datasets. The results in Table~\ref{ablation} provide important insights. Firstly, removing the background encoder (-background) results in a noticeable performance drop. This highlights the benefit of enhancing entity representations with background facts. Secondly, removing the support-specific adjustment module (-adjustment) leads to a significant decline in performance, emphasizing the crucial role of adjusting relation representations to capture meta relational information. Notably, -adjustment suffers the most significant performance degradation on the F-WikiPeople and F-WD50K datasets, which have a large amount of background data. This suggests that the richer the information in the generated relation representations, the more necessary it is to capture meta relational information.
\begin{figure}[t]
  \centering  
  \includegraphics[width=0.45\textwidth]{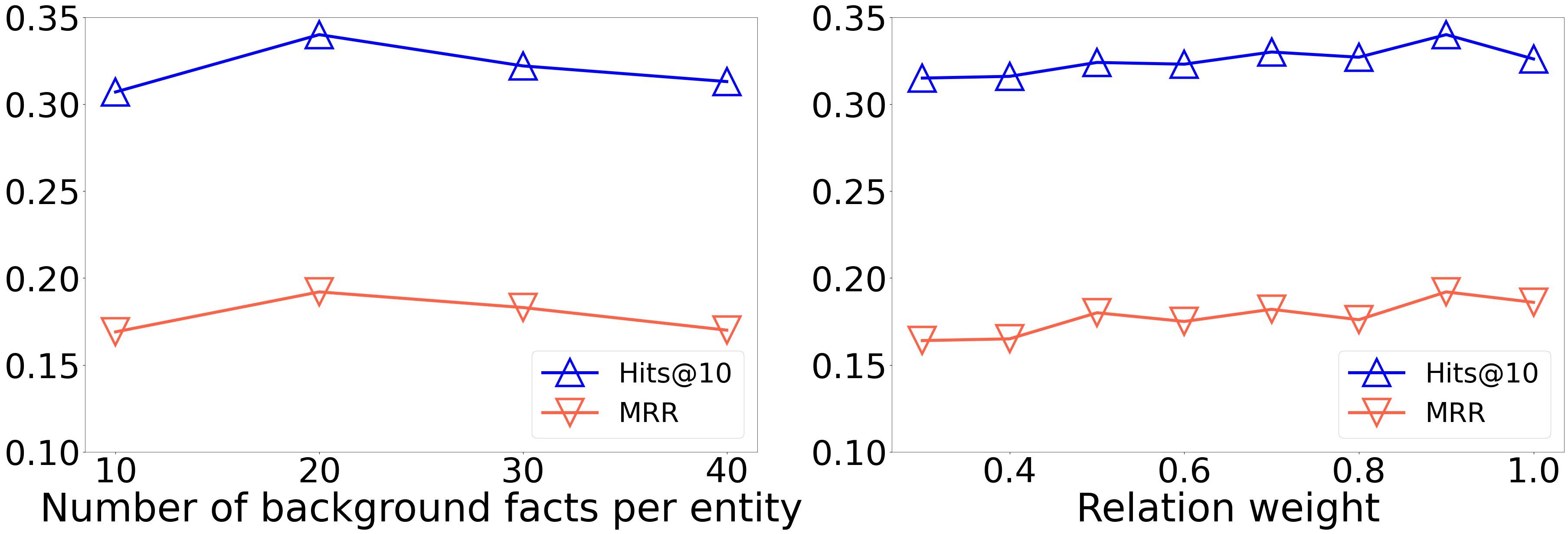}
  \caption{Impact of the number of background facts per entity and relation weight on the F-WD50K dataset.}
  \label{Fig.neighbor_weight}
\end{figure}

\subsection{Analysis on key Parameters}
The two key parameters of MetaRH are the maximum number of background facts per entity ($L$) and relation weight ($\tau$). To analyze the impact of these parameters on MetaRH's performance, experiments are conducted on the largest dataset F-WD50K. Figure~\ref{Fig.neighbor_weight} illustrates that $L$ significantly affects MetaRH's performance. If $L$ is too small, important background facts may be lost, while too large may result in insufficient attention to the most useful facts. In terms of $\tau$, the optimal performance of MetaRH is achieved when $\tau$ is set to 0.9, indicating that relations carry most of the type information of hyper-relational facts, compared to auxiliary attribute-value pairs, in the F-WD50K dataset.
    \begin{table}[t]\scriptsize
        \setlength{\tabcolsep}{1.25mm}{
        \begin{tabular}{l|c|c}
        \hline
        Query & MetaRH & GANA \\
        \hline
        \begin{tabular}[l]{@{}l@{}}
            ((Prince of Wales (Q180729),position held (P39),\\ \textcolor{red}{\text{monarch (Q116)}}),\\ \{(of (P642), Irish Free State (Q31747))\})
        \end{tabular} & 32 & 7 \\
        \hline
        \begin{tabular}[l]{@{}l@{}}
            ((Steve Jobs (Q19837),position held (P39),\\ \textcolor{red}{\text{chief executive officer (Q484876)}}),\\ \{(of (P642), Apple (Q312))\})
        \end{tabular} & 7 & 151 \\
        \hline
        \begin{tabular}[l]{@{}l@{}}
            ((Victor Hugo (Q535),position held (P39),\\ \textcolor{red}{\text{president (Q30461)}}),\\ \{(of (P642), Literary Society (Q3488144))\})
        \end{tabular} & 81 & 373 \\
        \hline
        \begin{tabular}[l]{@{}l@{}}
            (Second Punic War (Q6271),participant (P710),\\ \textcolor{red}{\text{Macedonia (Q83958)}}) \end{tabular} & 1 & 3 \\
        \hline
        \begin{tabular}[l]{@{}l@{}}
            (Rhine (Q584),basin country (P205),\\ \textcolor{red}{\text{Switzerland (Q39)}}) \end{tabular} & 8 & 8 \\
        \hline
        \begin{tabular}[l]{@{}l@{}}
            (Operation Barbarossa (Q83055),\\ participant (P710), \textcolor{red}{\text{Romania (Q203493)}}) \end{tabular} & 31 & 44 \\
        \hline
        \end{tabular}
        \caption{Case study on the F-WD50K dataset. Tail entities in these case facts are assumed to be predicted and are highlighted in red. The second and third columns are the ranks of correct answers for MetaRH and the best baseline GANA, respectively.}
        \label{Case}
        }
\end{table}
\subsection{Case Study}
To analyze the practical performance of MetaRH, we randomly selected 6 queries from the F-WD50K dataset for the case study, including 3 hyper-relational facts and 3 binary facts. As shown in Table~\ref{Case}, MetaRH outperforms GANA on most facts, demonstrating its superior performance. For the first case of answering the position held by the Prince of Wales in the Irish Free State, MetaRH does not perform as well as GANA. This can be attributed to a lack of background data. Since there are only three facts related to the Irish Free State in the background data, it prevents MetaRH from understanding the auxiliary attribute value pairs.

\section{Conclusion}
In this paper, we introduced a new task that is practical in real-world scenarios, called Few-Shot Link Prediction on Hyper-relational Facts (FSLPHFs). We defined the task and proposed a solution model called MetaRH, which consists of three modules: relation learning, support-specific adjustment, and query inference. These modules generate initial few-shot relation representations, adjust them based on the support set, and make inferences about queries, respectively. In addition, we constructed three datasets to test our approach. The experimental results show a significant improvement of MetaRH over existing models. In future research, we plan to utilize LLMs to reduce the dependence on background data and training tasks.

\section{Acknowledgments}
The work is supported by the National Natural Science Foundation of China under grant 62002341, 62306299, the National Key Research and Development Project of China, Beijing Academy of Artificial Intelligence under grant BAAI2019ZD0306, the KGJ Project under grant JCKY2022130C039, SMP-IDATA Open Youth Fund, and the Lenovo-CAS Joint Lab Youth Scientist Project. We thank anonymous reviewers for their insightful comments and suggestions.

\nocite{*}
\section{Bibliographical References}\label{sec:reference}
\bibliographystyle{lrec-coling2024-natbib}
\bibliography{lrec-coling2024-example}

\begin{thebibliography}{44}
\expandafter\ifx\csname natexlab\endcsname\relax\def\natexlab#1{#1}\fi

\bibitem[{Abboud et~al.(2020)Abboud, Ceylan, Lukasiewicz, and
  Salvatori}]{abboud2020boxe}
Ralph Abboud, Ismail Ceylan, Thomas Lukasiewicz, and Tommaso Salvatori. 2020.
\newblock Boxe: A box embedding model for knowledge base completion.
\newblock \emph{Advances in Neural Information Processing Systems},
  33:9649--9661.

\bibitem[{Bala{\v{z}}evi{\'c} et~al.(2019)Bala{\v{z}}evi{\'c}, Allen, and
  Hospedales}]{DBLP:conf/emnlp/BalazevicAH19}
Ivana Bala{\v{z}}evi{\'c}, Carl Allen, and Timothy~M Hospedales. 2019.
\newblock Tucker: Tensor factorization for knowledge graph completion.
\newblock \emph{arXiv preprint arXiv:1901.09590}.

\bibitem[{Bollacker et~al.(2008)Bollacker, Evans, Paritosh, Sturge, and
  Taylor}]{bollacker2008freebase}
Kurt Bollacker, Colin Evans, Praveen Paritosh, Tim Sturge, and Jamie Taylor.
  2008.
\newblock Freebase: a collaboratively created graph database for structuring
  human knowledge.
\newblock In \emph{Proceedings of the 2008 ACM SIGMOD international conference
  on Management of data}, pages 1247--1250.

\bibitem[{Bordes et~al.(2013)Bordes, Usunier, Garcia-Duran, Weston, and
  Yakhnenko}]{NIPS2013_1cecc7a7}
Antoine Bordes, Nicolas Usunier, Alberto Garcia-Duran, Jason Weston, and Oksana
  Yakhnenko. 2013.
\newblock Translating embeddings for modeling multi-relational data.
\newblock \emph{Advances in neural information processing systems}, 26.

\bibitem[{Brown et~al.(2020)Brown, Mann, Ryder, Subbiah, Kaplan, Dhariwal,
  Neelakantan, Shyam, Sastry, Askell et~al.}]{brown2020language}
Tom Brown, Benjamin Mann, Nick Ryder, Melanie Subbiah, Jared~D Kaplan, Prafulla
  Dhariwal, Arvind Neelakantan, Pranav Shyam, Girish Sastry, Amanda Askell,
  et~al. 2020.
\newblock Language models are few-shot learners.
\newblock \emph{Advances in neural information processing systems},
  33:1877--1901.

\bibitem[{Chen et~al.(2019)Chen, Zhang, Zhang, Chen, and
  Chen}]{DBLP:conf/emnlp/ChenZZCC19}
Mingyang Chen, Wen Zhang, Wei Zhang, Qiang Chen, and Huajun Chen. 2019.
\newblock Meta relational learning for few-shot link prediction in knowledge
  graphs.
\newblock \emph{arXiv preprint arXiv:1909.01515}.

\bibitem[{Codd(1983)}]{Codd}
Edgar~Frank Codd. 1983.
\newblock A relational model of data for large shared data banks.
\newblock \emph{Communications of the ACM}, 26(1):64--69.

\bibitem[{Di et~al.(2021)Di, Yao, and Chen}]{DBLP:conf/www/DiYC21}
Shimin Di, Quanming Yao, and Lei Chen. 2021.
\newblock Searching to sparsify tensor decomposition for n-ary relational data.
\newblock In \emph{Proceedings of the Web Conference 2021}, pages 4043--4054.

\bibitem[{Dong et~al.(2015)Dong, Wei, Zhou, and Xu}]{DBLP:conf/acl/DongWZX15}
Li~Dong, Furu Wei, Ming Zhou, and Ke~Xu. 2015.
\newblock Question answering over freebase with multi-column convolutional
  neural networks.
\newblock In \emph{Proceedings of the 53rd Annual Meeting of the Association
  for Computational Linguistics and the 7th International Joint Conference on
  Natural Language Processing (Volume 1: Long Papers)}, pages 260--269.

\bibitem[{Fatemi et~al.(2019)Fatemi, Taslakian, Vazquez, and
  Poole}]{DBLP:journals/corr/abs-1906-00137}
Bahare Fatemi, Perouz Taslakian, David Vazquez, and David Poole. 2019.
\newblock Knowledge hypergraphs: Extending knowledge graphs beyond binary
  relations.
\newblock \emph{arXiv preprint arXiv:1906.00137}.

\bibitem[{Finn et~al.(2017)Finn, Abbeel, and Levine}]{DBLP:conf/icml/FinnAL17}
Chelsea Finn, Pieter Abbeel, and Sergey Levine. 2017.
\newblock Model-agnostic meta-learning for fast adaptation of deep networks.
\newblock In \emph{International conference on machine learning}, pages
  1126--1135.

\bibitem[{Galkin et~al.(2020)Galkin, Trivedi, Maheshwari, Usbeck, and
  Lehmann}]{DBLP:conf/emnlp/GalkinTMUL20}
Mikhail Galkin, Priyansh Trivedi, Gaurav Maheshwari, Ricardo Usbeck, and Jens
  Lehmann. 2020.
\newblock Message passing for hyper-relational knowledge graphs.
\newblock \emph{arXiv preprint arXiv:2009.10847}.

\bibitem[{Guan et~al.(2020)Guan, Jin, Guo, Wang, and
  Cheng}]{DBLP:conf/acl/GuanJGWC20}
Saiping Guan, Xiaolong Jin, Jiafeng Guo, Yuanzhuo Wang, and Xueqi Cheng. 2020.
\newblock Neuinfer: Knowledge inference on n-ary facts.
\newblock In \emph{Proceedings of the 58th annual meeting of the association
  for computational linguistics}, pages 6141--6151.

\bibitem[{Guan et~al.(2021)Guan, Jin, Guo, Wang, and Cheng}]{Guan2021LinkPO}
Saiping Guan, Xiaolong Jin, Jiafeng Guo, Yuanzhuo Wang, and Xueqi Cheng. 2021.
\newblock Link prediction on n-ary relational data based on relatedness
  evaluation.
\newblock \emph{IEEE Transactions on Knowledge and Data Engineering},
  35(1):672--685.

\bibitem[{Guan et~al.(2018)Guan, Jin, Wang, and
  Cheng}]{DBLP:conf/cikm/GuanJWC18}
Saiping Guan, Xiaolong Jin, Yuanzhuo Wang, and Xueqi Cheng. 2018.
\newblock Shared embedding based neural networks for knowledge graph
  completion.
\newblock In \emph{Proceedings of the 27th ACM International Conference on
  Information and Knowledge Management}, pages 247--256.

\bibitem[{Guan et~al.(2019)Guan, Jin, Wang, and
  Cheng}]{DBLP:conf/www/GuanJWC19}
Saiping Guan, Xiaolong Jin, Yuanzhuo Wang, and Xueqi Cheng. 2019.
\newblock Link prediction on n-ary relational data.
\newblock In \emph{The world wide web conference}, pages 583--593.

\bibitem[{Jiang et~al.(2021)Jiang, Gao, and Lv}]{DBLP:conf/sigir/JiangGL21}
Zhiyi Jiang, Jianliang Gao, and Xinqi Lv. 2021.
\newblock Metap: Meta pattern learning for one-shot knowledge graph completion.
\newblock In \emph{Proceedings of the 44th International ACM SIGIR Conference
  on Research and Development in Information Retrieval}, pages 2232--2236.

\bibitem[{Kazemi and Poole(2018)}]{DBLP:conf/nips/Kazemi018}
Seyed~Mehran Kazemi and David Poole. 2018.
\newblock Simple embedding for link prediction in knowledge graphs.
\newblock \emph{Advances in neural information processing systems}, 31.

\bibitem[{Kingma and Ba(2014)}]{kingma2015adam}
Diederik~P Kingma and Jimmy Ba. 2014.
\newblock Adam: A method for stochastic optimization.
\newblock \emph{arXiv preprint arXiv:1412.6980}.

\bibitem[{Krieger and Willms(2015)}]{krieger2015extending}
Hans-Ulrich Krieger and Christian Willms. 2015.
\newblock Extending owl ontologies by cartesian types to represent n-ary
  relations in natural language.
\newblock In \emph{Proceedings of the 1st Workshop on Language and Ontologies}.

\bibitem[{Liu et~al.(2020{\natexlab{a}})Liu, Yao, and
  Li}]{DBLP:conf/www/LiuY20}
Yu~Liu, Quanming Yao, and Yong Li. 2020{\natexlab{a}}.
\newblock Generalizing tensor decomposition for n-ary relational knowledge
  bases.
\newblock In \emph{Proceedings of the web conference 2020}, pages 1104--1114.

\bibitem[{Liu et~al.(2020{\natexlab{b}})Liu, Yao, and
  Li}]{DBLP:conf/www/0016Y020}
Yu~Liu, Quanming Yao, and Yong Li. 2020{\natexlab{b}}.
\newblock Generalizing tensor decomposition for n-ary relational knowledge
  bases.
\newblock In \emph{Proceedings of the web conference 2020}, pages 1104--1114.

\bibitem[{Lukovnikov et~al.(2017)Lukovnikov, Fischer, Lehmann, and
  Auer}]{DBLP:conf/www/LukovnikovFLA17}
Denis Lukovnikov, Asja Fischer, Jens Lehmann, and S{\"o}ren Auer. 2017.
\newblock Neural network-based question answering over knowledge graphs on word
  and character level.
\newblock In \emph{Proceedings of the 26th international conference on World
  Wide Web}, pages 1211--1220.

\bibitem[{Luo et~al.(2023)Luo, Yang, Guo, Sun, Yao, Tang, Wan, Song, Lin
  et~al.}]{luo2023hahe}
Haoran Luo, Yuhao Yang, Yikai Guo, Mingzhi Sun, Tianyu Yao, Zichen Tang,
  Kaiyang Wan, Meina Song, Wei Lin, et~al. 2023.
\newblock Hahe: Hierarchical attention for hyper-relational knowledge graphs in
  global and local level.
\newblock \emph{arXiv preprint arXiv:2305.06588}.

\bibitem[{Maas et~al.(2013)Maas, Hannun, Ng et~al.}]{maas2013rectifier}
Andrew~L Maas, Awni~Y Hannun, Andrew~Y Ng, et~al. 2013.
\newblock Rectifier nonlinearities improve neural network acoustic models.
\newblock In \emph{Proc. icml}, volume~30, page~3. Atlanta, GA.

\bibitem[{Munkhdalai and Yu(2017)}]{metanetwork}
Tsendsuren Munkhdalai and Hong Yu. 2017.
\newblock Meta networks.
\newblock In \emph{International conference on machine learning}, pages
  2554--2563.

\bibitem[{Nguyen et~al.(2014)Nguyen, Bodenreider, and Sheth}]{nguyen2014don}
Vinh Nguyen, Olivier Bodenreider, and Amit Sheth. 2014.
\newblock Don't like rdf reification? making statements about statements using
  singleton property.
\newblock In \emph{Proceedings of the 23rd international conference on World
  wide web}, pages 759--770.

\bibitem[{Niu et~al.(2021)Niu, Li, Tang, Geng, Dai, Liu, Wang, Sun, Huang, and
  Si}]{DBLP:conf/sigir/NiuLTGDLWSHS21}
Guanglin Niu, Yang Li, Chengguang Tang, Ruiying Geng, Jian Dai, Qiao Liu, Hao
  Wang, Jian Sun, Fei Huang, and Luo Si. 2021.
\newblock Relational learning with gated and attentive neighbor aggregator for
  few-shot knowledge graph completion.
\newblock In \emph{Proceedings of the 44th International ACM SIGIR conference
  on research and development in information retrieval}, pages 213--222.

\bibitem[{Pan et~al.(2024)Pan, Luo, Wang, Chen, Wang, and Wu}]{pan2024unifying}
Shirui Pan, Linhao Luo, Yufei Wang, Chen Chen, Jiapu Wang, and Xindong Wu.
  2024.
\newblock Unifying large language models and knowledge graphs: A roadmap.
\newblock \emph{IEEE Transactions on Knowledge and Data Engineering}.

\bibitem[{Rosso et~al.(2020)Rosso, Yang, and
  Cudr{\'e}-Mauroux}]{DBLP:conf/www/RossoYC20}
Paolo Rosso, Dingqi Yang, and Philippe Cudr{\'e}-Mauroux. 2020.
\newblock Beyond triplets: hyper-relational knowledge graph embedding for link
  prediction.
\newblock In \emph{Proceedings of the web conference 2020}, pages 1885--1896.

\bibitem[{Sheng et~al.(2020)Sheng, Guo, Chen, Yue, Wang, Liu, and
  Xu}]{DBLP:conf/emnlp/ShengGCYWLX20}
Jiawei Sheng, Shu Guo, Zhenyu Chen, Juwei Yue, Lihong Wang, Tingwen Liu, and
  Hongbo Xu. 2020.
\newblock Adaptive attentional network for few-shot knowledge graph completion.
\newblock \emph{arXiv preprint arXiv:2010.09638}.

\bibitem[{Sun et~al.(2019)Sun, Deng, Nie, and Tang}]{sunrotate}
Zhiqing Sun, Zhi-Hong Deng, Jian-Yun Nie, and Jian Tang. 2019.
\newblock Rotate: Knowledge graph embedding by relational rotation in complex
  space.
\newblock \emph{arXiv preprint arXiv:1902.10197}.

\bibitem[{Vrande{\v{c}}i{\'c} and Kr{\"o}tzsch(2014)}]{vrandevcic2014wikidata}
Denny Vrande{\v{c}}i{\'c} and Markus Kr{\"o}tzsch. 2014.
\newblock Wikidata: a free collaborative knowledgebase.
\newblock \emph{Communications of the ACM}, 57(10):78--85.

\bibitem[{Wang et~al.(2023)Wang, Wang, Li, Chen, and Li}]{wang2023hyconve}
Chenxu Wang, Xin Wang, Zhao Li, Zirui Chen, and Jianxin Li. 2023.
\newblock Hyconve: A novel embedding model for knowledge hypergraph link
  prediction with convolutional neural networks.
\newblock In \emph{Proceedings of the ACM Web Conference 2023}, pages 188--198.

\bibitem[{Wang et~al.(2017)Wang, Mao, Wang, and Guo}]{wang2017knowledge}
Quan Wang, Zhendong Mao, Bin Wang, and Li~Guo. 2017.
\newblock Knowledge graph embedding: A survey of approaches and applications.
\newblock \emph{IEEE Transactions on Knowledge and Data Engineering},
  29(12):2724--2743.

\bibitem[{Wang et~al.(2021)Wang, Wang, Lyu, and Zhu}]{DBLP:conf/acl/WangWLZ21}
Quan Wang, Haifeng Wang, Yajuan Lyu, and Yong Zhu. 2021.
\newblock Link prediction on n-ary relational facts: A graph-based approach.
\newblock \emph{arXiv preprint arXiv:2105.08476}.

\bibitem[{Wang et~al.(2014)Wang, Zhang, Feng, and
  Chen}]{DBLP:conf/aaai/WangZFC14}
Zhen Wang, Jianwen Zhang, Jianlin Feng, and Zheng Chen. 2014.
\newblock Knowledge graph embedding by translating on hyperplanes.
\newblock In \emph{Proceedings of the AAAI conference on artificial
  intelligence}, volume~28.

\bibitem[{Wen et~al.(2016)Wen, Li, Mao, Chen, and
  Zhang}]{DBLP:conf/ijcai/WenLMCZ16}
Jianfeng Wen, Jianxin Li, Yongyi Mao, Shini Chen, and Richong Zhang. 2016.
\newblock On the representation and embedding of knowledge bases beyond binary
  relations.
\newblock \emph{arXiv preprint arXiv:1604.08642}.

\bibitem[{Xiong et~al.(2023)Xiong, Nayyer, Pan, and Staab}]{xiong2023shrinking}
Bo~Xiong, Mojtaba Nayyer, Shirui Pan, and Steffen Staab. 2023.
\newblock Shrinking embeddings for hyper-relational knowledge graphs.
\newblock \emph{arXiv preprint arXiv:2306.02199}.

\bibitem[{Xiong et~al.(2018)Xiong, Yu, Chang, Guo, and
  Wang}]{DBLP:conf/emnlp/XiongYCGW18}
Wenhan Xiong, Mo~Yu, Shiyu Chang, Xiaoxiao Guo, and William~Yang Wang. 2018.
\newblock One-shot relational learning for knowledge graphs.
\newblock \emph{arXiv preprint arXiv:1808.09040}.

\bibitem[{Yan et~al.(2022)Yan, Zhang, Sun, Xu, Li, Liu, Liu, and
  Wang}]{DBLP:conf/aaai/YanSY2022}
Shiyao Yan, Zequn Zhang, Xian Sun, Guangluan Xu, Shuchao Li, Qing Liu, Nayu
  Liu, and Shensi Wang. 2022.
\newblock Polygone: Modeling n-ary relational data as gyro-polygons in
  hyperbolic space.
\newblock In \emph{Proceedings of the AAAI Conference on Artificial
  Intelligence}, volume~36, pages 4308--4317.

\bibitem[{Zhang et~al.(2020)Zhang, Yao, Huang, Jiang, Li, and
  Chawla}]{DBLP:conf/aaai/ZhangYHJLC20}
Chuxu Zhang, Huaxiu Yao, Chao Huang, Meng Jiang, Zhenhui Li, and Nitesh~V
  Chawla. 2020.
\newblock Few-shot knowledge graph completion.
\newblock In \emph{Proceedings of the AAAI conference on artificial
  intelligence}.

\bibitem[{Zhang et~al.(2018)Zhang, Li, Mei, and Mao}]{DBLP:conf/www/ZhangLMM18}
Richong Zhang, Junpeng Li, Jiajie Mei, and Yongyi Mao. 2018.
\newblock Scalable instance reconstruction in knowledge bases via relatedness
  affiliated embedding.
\newblock In \emph{Proceedings of the 2018 world wide web conference}, pages
  1185--1194.

\bibitem[{Zhu et~al.(2023)Zhu, Wang, Chen, Qiao, Ou, Yao, Deng, Chen, and
  Zhang}]{zhu2023llms}
Yuqi Zhu, Xiaohan Wang, Jing Chen, Shuofei Qiao, Yixin Ou, Yunzhi Yao, Shumin
  Deng, Huajun Chen, and Ningyu Zhang. 2023.
\newblock Llms for knowledge graph construction and reasoning: Recent
  capabilities and future opportunities.
\newblock \emph{arXiv preprint arXiv:2305.13168}.

\end{thebibliography}

\label{lr:ref}
\bibliographystylelanguageresource{lrec-coling2024-natbib}
\bibliographylanguageresource{languageresource}

\end{document}